\documentclass[10pt,twocolumn,letterpaper]{article}

\usepackage{iccv}
\usepackage{times}
\usepackage{epsfig}
\usepackage{graphicx}
\usepackage{amsmath}
\usepackage{amssymb}
\usepackage{bm}
\usepackage{subfigure}
\usepackage{booktabs}

\usepackage[pagebackref=true,breaklinks=true,letterpaper=true,colorlinks,bookmarks=false]{hyperref}

\iccvfinalcopy 

\def\iccvPaperID{10968} 

\ificcvfinal\fi

\begin{document}

\title{Diffusion-based Target Sampler for Unsupervised Domain Adaptation}

\author{Yulong Zhang$^1$\thanks{Equal contribution.} \quad Shuhao Chen$^2$\footnotemark[1] \quad Yu Zhang$^{2}$\thanks{Corresponding author.} \quad Jiangang Lu$^1$\\
$^1$Zhejiang University\\
$^2$Southern University of Science and Technology\\
\!\!\!{\tt\small zhangylcse@zju.edu.cn, 12232388@mail.sustech.edu.cn, yu.zhang.ust@gmail.com, lujg@zju.edu.cn}
}

\maketitle
\ificcvfinal\thispagestyle{empty}\fi

\begin{abstract}
Limited transferability hinders the performance of deep learning models when applied to new application scenarios.
Recently, unsupervised domain adaptation (UDA) has achieved significant progress in addressing this issue via learning domain-invariant features.
However, large domain shifts and the sample scarcity in the target domain make existing UDA methods achieve suboptimal performance.
To alleviate these issues, we propose a plug-and-play Diffusion-based Target Sampler (DTS) to generate high fidelity and diversity pseudo target samples.
By introducing class-conditional information, the labels of the generated target samples can be controlled.
The generated samples can well simulate the data distribution of the target domain and help existing UDA methods transfer from the source domain to the target domain more easily, thus improving the transfer performance.
Extensive experiments on various benchmarks demonstrate that the performance of existing UDA methods can be greatly improved through the proposed DTS method.
The code will be released soon.
\end{abstract}

\section{Introduction}
With the development of data collection and computing facilities, deep learning models have achieved remarkable advances in a variety of applications for their powerful representation learning capabilities~\cite{wang2023energyinspired, Shi_2021_ICCV, subramanian2022generalization}.
However, when domain shifts occur, the well-trained models in a source domain will suffer significant performance degradation in a target domain.
In this case, large-scale target samples need to be recollected and labeled, which is time-consuming and expensive.

Unsupervised Domain Adaptation (UDA) \cite{yang2020transfer} leverages transferable knowledge from the source domain and applies it to an unlabeled target domain to improve the reusability of existing models and data \cite{sun2022prior, goyal2022test, jiang2022transferability}.
To alleviate domain shifts, existing UDA methods mainly learn a domain-invariant feature representation explicitly or implicitly.
Specifically, some UDA methods explicitly  minimize the distribution discrepancy based on distance measures between the source and target domains \cite{ouyang2021maximum, shen2018wasserstein, NEURIPS2021_ae0909a3}.
Inspired by generative adversarial networks (GAN) \cite{goodfellow2014generative}, a domain discriminator is introduced to bridge the source and target domains implicitly through adversarial training \cite{zhang2022spectral, chen2022reusing, rangwani2022closer}.
However, under the UDA setting where the domain shift occurs, the limited or even scarce target samples cannot represent the data distribution in the target domain accurately, which limits the performance of existing UDA methods.
In other words, direct domain alignments based on source samples and scarce unlabeled target samples will lead to suboptimal transfer effects.

To address the aforementioned limitations, in this paper, we propose a Diffusion-based Target Sampler (DTS) for UDA.  The proposed DTS is to generate pseudo target samples that could follow the target distribution.
In this way, target samples can be augmented with pseudo target samples and they could improve the performance of UDA models.
Specifically, the DTS method adopts the Diffusion Probabilistic Model (DPM) \cite{ho2020denoising, dhariwal2021diffusion}, which has recently been proposed to achieve superior performance in image generation, to generate high-quality target samples.
Based on pseudo labels given by existing UDA methods for target samples, the DPM is trained conditionally to achieve class-conditional generation, and hence the pseudo labels of generated samples can be naturally determined.
Then generated target samples and original source samples are combined as the augmented source domain, where original source samples are used to suppress the effect of noisy labels of generated target samples.
In this manner, the distribution of the augmented source domain is closer to the target domain, which reduces the difficulty of domain adaptation (DA).
Note that the proposed DTS framework is a plug-and-play module that can be embedded into any existing UDA methods to improve their transfer performance, which has rarely been investigated in the literature.
We combine the proposed DTS framework with state-of-the-art UDA methods and conduct extensive experiments on three UDA benchmark datasets to demonstrate the superiority of the proposed DTS framework.

The contributions of this paper are three-fold.
\begin{itemize}
\item  We propose a novel DTS method that directly generates high fidelity and diversity samples that are likely to follow the distribution of the target domain. To the best of our knowledge, this is the first application of DPMs in UDA.
\item The proposed plug-and-play DTS framework can be embedded into any existing UDA method to improve their transfer performance.
\item Extensive experiments on UDA benchmark datasets  demonstrate the effectiveness of the proposed DTS framework. The augmented source domain consisting of source samples and generated target samples can help reduce the domain discrepancy.
\end{itemize}

\section{Related Work} \label{section:RW}
\subsection{Unsupervised Domain Adaptation}
UDA methods \cite{zhao2020review, long2015learning, ganin2016domain} extract knowledge from the labeled source domain to facilitate learning in the unlabeled target domain.
As the distribution of the target domain differs from that of the source domain, various methods are proposed to reduce the domain discrepancy and they can be mainly classified into two categories: discrepancy-based methods and adversarial-based methods.

Discrepancy-based methods learn a feature extractor to minimize the distribution discrepancy between the source and target domains.
For example, \cite{long2015learning, zhu2020deep} try to minimize the maximum mean discrepancy (MMD) \cite{gretton2012kernel} across domains, while MCC \cite{jin2020minimum} minimizes the class confusion loss without explicit domain alignment.
Margin disparity discrepancy (MDD) \cite{MDD} is to perform domain alignment with analyses in terms of generalization bounds.
\cite{shen2018wasserstein} uses the Wasserstein distance for distributional alignment to achieve better transfer effects.

On the other hand, adversarial-based methods align distributions across domains through adversarial training \cite{goodfellow2014explaining}.
For example, DANN \cite{ganin2016domain} adversarially learns a domain discriminator to distinguish samples in two domains and thus enables the feature extractor to confuse the domain discriminator. 
CDAN \cite{long2018conditional} injects class-specific information into the discriminator to facilitate the alignment of multi-modal distributions.
SDAT \cite{rangwani2022closer} uses sharpness-aware minimization \cite{foret2020sharpness} to seek a flat minimize for better generalization, while ELS \cite{zhang2023free} improves over SDAT by introducing label smoothing to domain labels and achieves better transfer performance.

Adversarial training can also be used to generate samples to bridge the source and target domains.
CyCADA \cite{hoffman2018cycada} adopts GAN for image-to-image translation via the cycle consistency loss \cite{zhu2017unpaired} and semantic consistency loss.
Specifically, it directly changes the style of source images to that of the target domain to obtain a better classifier for the target domain.
\cite{yang2020bi} uses two cross-domain generators to synthesize data of each domain  conditioned on the other, and learns two task-specific classifiers.
\cite{gao2022middlegan} bridges the source and target domains by generating the intermediate domain which is similar to both domains.
Compared with those methods that generate an intermediate domain to interpolate between the distributions of the source and target domains, the proposed DTS framework directly generates pseudo target samples that obey the target distribution.
Instead of using adversarial training strategies, the proposed DTS framework is based on the DPM, which has better generation capabilities and is easier to converge in the training process \cite{dhariwal2021diffusion}.
Moreover, the generated categories and the number of generated samples can be flexibly controlled by DTS, which is more advantageous than the UDA methods based on image-to-image translation.

\subsection{Diffusion Probabilistic Model}
DPM \cite{sohl2015deep} has achieved great success in various generative tasks in recent years.
Acting as the Markov chain, DPM includes a forward diffusion process and a reverse denoising process, which convert the original data into Gaussian noise and deduce the Gaussian noise into the original data gradually.
Specifically, the forward process gradually adds Gaussian noise to the input image $\bm{x}_0$ until the image is completely converted to a random noise $\bm{x}_T$,
i.e., $q\left( {{\bm{x}_t}| {{\bm{x}_{t - 1}}}} \right) = \mathcal{N} \left( {\sqrt {1 - {\beta _t}} {\bm{x}_{t - 1}},{\beta _t}\bf{I} } \right)$, where $\mathcal{N}(\cdot,\cdot)$ denotes a normal distribution with mean and variance specified by arguments, and ${\beta _t}$ increases gradually according to a variance schedule.
The noise obtained in the forward process can be regarded as a series of labels which the model ${\epsilon _\theta }\left( {{\bm{x}_t},t} \right)$ is trained to predict in the reverse process, where $\theta$ denotes the parameters of the model.
In practice, the reverse process can be formulated as optimizing a variational upper bound on the negative log likelihood, which can be formalized as the Kullback-Leibler (KL) divergence \cite{ho2020denoising, song2020denoising, nichol2021improved,huang2021variational}.
After the training process, the model can gradually denoise a random noise $\bm{x}_T$ to a generated image $\bm{x}_0$ according to ${\bm{x}_{t - 1}} = \frac{1}{{\sqrt {{\alpha _t}} }}\left( {{\bm{x}_t} - \frac{{1 - {\alpha _t}}}{{\sqrt {1 - {{\bar \alpha }_t}} }}{{\rm{\epsilon}}_\theta }\left( {{\bm{x}_t},t} \right)} \right) + {\sigma _t}z$, where $z \sim \mathcal{N}\left( {0,\bf{I}} \right)$, ${\alpha _t} = 1 - {\beta _t}$, ${{\bar \alpha }_t} = \prod\nolimits_{s = 0}^t {{\alpha _s}} $, and ${\sigma _t} = \frac{{1 - {{\bar \alpha }_{t - 1}}}}{{1 - {{\bar \alpha }_t}}}{\beta _t}$.


\begin{figure*}[t]
\centering
\includegraphics[width=6.5in]{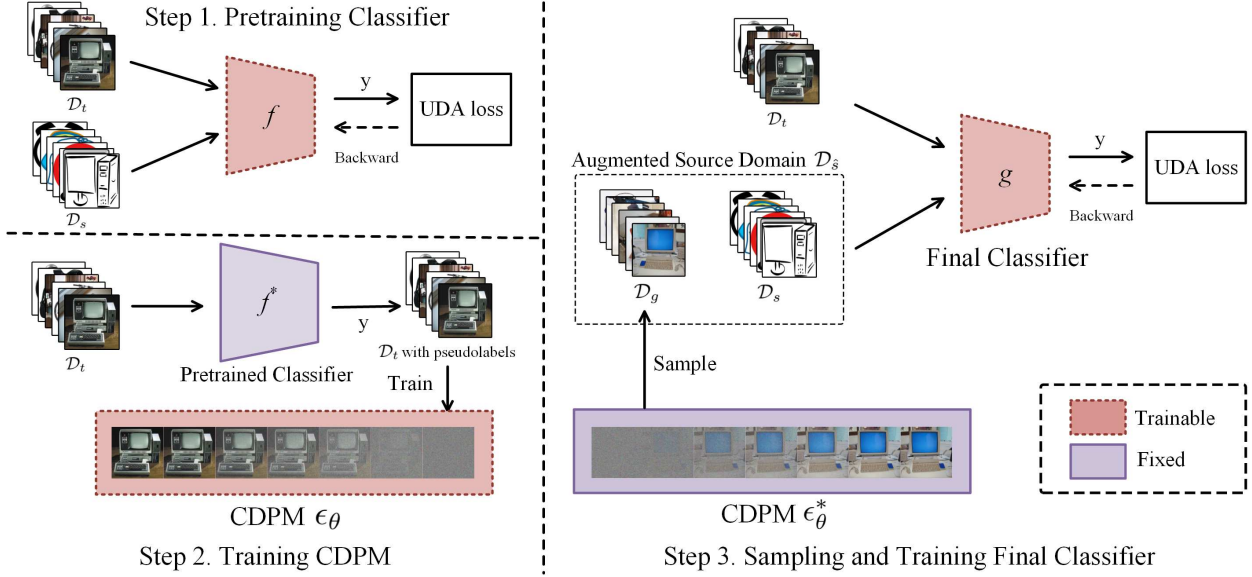}
\caption{An overview of the proposed DTS framework. In step 1, we pretrain the target classifier with some existing UDA methods. In step 2, with the pretrained classifier, we obtain the pseudo labels of target domain samples to train CDPM $\epsilon _\theta$. In step 3, the generated target domain $\mathcal{D}_g$ can be obtained through sampling the pretrained CDPM $\epsilon _\theta^{*}$ in step 2. Finally, the final classifier is trained by transferring from the augmented source domain ${\mathcal{D}_{\hat{s}}}$ to the target domain $\mathcal{D}_t$.}
\label{model}
\end{figure*}

Although DPM has obtained superior performance in image generation, it still has the problem of slow sampling speed \cite{nichol2021improved, lu2022dpmsolver} due to thousands of denoising steps required to generate a sample of high quality, which greatly hinders the application of DPM \cite{ho2020denoising, lu2022dpm, yang2022diffusion}.
To address this issue, denoising diffusion implicit model (DDIM) \cite{ho2020denoising} achieves 10$\times$ to 50$\times$ sampling acceleration through non-Markov diffusion processes.
DPM-Solver \cite{lu2022dpmsolver} formulates DPMs as diffusion ordinary differential equations to reduce the number of denoising steps from 1000 to about 10.
Moreover, \cite{lu2022dpm} proposes DPM-Solver++ to alleviate the problem of unstable guided sampling in DPM-Solver.
In this paper, to generate class-conditional samples, we use conditional diffusion probabilistic model (CDPM) \cite{nichol2021improved} in the DTS framework. 
For fast sampling, we use DPM-Solver++ to accelerate the sampling process.

\section{The DTS Method}

In UDA, we are given a labeled source domain ${\mathcal{D}_s} = \{ (\bm{x}_i^{sd},y_i^{sd})\} _{i = 1}^{{N_s}}$ and an unlabeled target domain ${\mathcal{D}_t} = \{\bm{x}_j^{td}\} _{j = 1}^{{N_t}}$.
We aim to explore useful information in $\mathcal{D}_s$ to help the prediction in $\mathcal{D}_t$. Usually $N_s$ is assumed to be much larger than $N_t$ and $N_t$ is not so large, making $\mathcal{D}_t$ not enough to model the entire data distribution of the target domain. To alleviate this problem, the proposed DTS method is to generate more target samples via DPM.

As illustrated in Figure~\ref{model}, the overall DTS framework can be divided into three steps.
In step 1, we use some existing UDA methods to obtain the target classifier. In step 2, pseudo labels of target samples are assigned by the pretrained classifier in step 1 and target samples with pseudo labels are used to train the CDPM $\epsilon_\theta$.
In step 3, the pretrained CDPM $\epsilon_\theta^*$ is adopted to generate target samples and those generated target samples are combined with the original source samples as the augmented source domain ${\mathcal{D}_{\hat{s}}}$. The final classifier can be obtained by transferring from the augmented source domain ${\mathcal{D}_{\hat{s}}}$ to the target domain $\mathcal{D}_t$.
In the following sections, we will discuss each step in detail.

\subsection{Pretraining a Target Classifier}
The labels of target samples are not available under the UDA setting.
Therefore, to obtain pseudo labels of target samples, the proposed DTS framework adopts an existing UDA model to pretrain the target classifier.
Here step 1 starts by training a UDA model, whose objective function is formulated as
\begin{align}
\label{eqn1}
\!\!\!\!{f^*} \!\!=\!  \mathop {\arg \min }_{f}& \frac{1}{{{N_s}}}\!\!\sum\limits_{i = 1}^{{N_s}} \!\mathcal{L}_c (\! {f( {{\bm{x}_i^{sd}}} ),{y_i^{sd}}}\! )
\!\!+\!\!\lambda {{R(T(\mathcal{D}_s),\!T(\mathcal{D}_t)\!)}},\!\!\!
\end{align}
where $T$ denotes a feature transformation network on the source and target domains, $f(\cdot)$ denotes the classifier,
$\mathcal{L}_c(\cdot,\cdot)$ denotes the task loss in the source domain such as the cross-entropy loss, and $R(\cdot,\cdot)$ denotes a transfer regularization term such as the discrepancy-based loss \cite{ouyang2021maximum, shen2018wasserstein, NEURIPS2021_ae0909a3} and adversarial-based loss \cite{zhang2022spectral, chen2022reusing, rangwani2022closer}. 
After solving problem (\ref{eqn1}), we can obtain the pretrained classifier $f^*$.


\subsection{Training CDPM}
In step 2, we aim to simulate the target distribution via $P_\theta (\bm{x}^{td})$ with parameter $\theta$ to generate samples similar to the target domain.
Recall that it is not easy to transform noise to structured data, but it is much easier to convert structured data to noise.
In particular, we can use a forward process to convert an original image $\bm{x}_0^{td}$ into a noise $\bm{x}_T^{td}$ through $T$ time steps of a stochastic encoder $q\left( {{\bm{x}_t^{td}}| {{\bm{x}_{t - 1}^{td}}}} \right)$. Then we learn a reverse process to undo this process with a decoder ${{p_\theta }\left( {{\bm{x}_{t - 1}^{td}}| {{\bm{x}^{td}_t}}} \right)}$.
To achieve this, the mainstream method is to maximize the $\ln P_\theta (\bm{x}^{td})$ \cite{su2018variational}.
To construct an upper bound of the log likelihood, we add a KL divergence to the negative log likelihood function, where for brevity the superscript $td$ of $\bm{x}^{td}$ is omitted in the following formulas, as
\begin{align}
- \log {p_\theta }\left( {{\bm{x}_0}} \right) &\le  - \log {p_\theta }\left( {{\bm{x}_0}} \right) + {D_{KL}}\left( {q\left( {{\bm{x}_{1:T}}| {{\bm{x}_0}}} \right)} \right) \nonumber\\
&= {\mathbb{E}_{q\left( {{\bm{x}_{1:T}}| {{\bm{x}_0}}} \right)}}\left[ {\log \frac{{q\left( {{\bm{x}_{1:T}}|{{\bm{x}_0}}} \right)}}{{{p_\theta }\left( {{\bm{x}_{0:T}}} \right)}}} \right] \nonumber\\
&\buildrel \Delta \over = {L_{\textrm{VLB}}},\label{eqn5}
\end{align}
where $\bm{x}_0$ denotes the original target samples, and $D_{KL}(\cdot)$ denotes the KL divergence.
Built upon the Markov property and the Bayes rule, $L_{\textrm{VLB}}$ can be simplified as
\begin{align}
{L_{\textrm{VLB}}}& = {\mathbb{E}_{q\left( {{\bm{x}_{1:T}}|{{\bm{x}_0}}} \right)}} {\log \frac{{\prod\limits_{t = 1}^T {q\left( {{\bm{x}_t}| {{\bm{x}_{t - 1}}}} \right)} }}{{{p_\theta }\left( {{\bm{x}_T}} \right)\prod\limits_{t = 1}^T {{p_\theta }\left( {{\bm{x}_{t - 1}}| {{\bm{x}_t}}} \right)} }}}  \nonumber\\
&= {\mathbb{E}_{q\left( {{\bm{x}_{1:T}}|{{\bm{x}_0}}} \right)}}[{\underbrace {{D_{KL}}\left( {q\left( {{\bm{x}_T}|{{\bm{x}_0}}} \right) \| \;{{p_\theta }\left( {{\bm{x}_T}} \right)}} \right)}_{{L_T}}} \nonumber\\
&\ \ \ \ + {\sum\limits_{t = 2}^T {\underbrace {{D_{KL}}\left( {q\left( {{\bm{x}_{t - 1}}| {{\bm{x}_t},{\bm{x}_0}}} \right)\|\;{{p_\theta }\left( {{\bm{x}_{t - 1}}|{{\bm{x}_t}}} \right)}} \right)}_{{L_{t - 1}}}} }\nonumber \\
&\ \ \ \ \ {\underbrace { - \log {p_\theta }\left( {{\bm{x}_0}|{{\bm{x}_1}}} \right)}_{{L_0}}}],\label{eqn6}
\end{align}
where $L_T$ and $L_0$ terms can be easily computed  \cite{ho2020denoising}. $L_{t-1}$ can be simplified as
\begin{align}
{L_{t - 1}} &= {\mathbb{E}_{\epsilon \sim \mathcal{N}\left( {0,\bf{I}} \right)}}\left[ {{\lambda _t}{{\left\| {\epsilon  - {\epsilon _\theta }\left( {{\bm{x}_t},t} \right)} \right\|}^2}} \right]+C',\label{eqn7}
\end{align}
where $C'$ denotes a constant independent of $\theta$, ${\lambda _t} = \beta _t^2/2\sigma _t^2{\alpha _t}\left( {1 - {{\bar \alpha }_t}} \right)$, and ${\bm{x}_t} = \sqrt {{{\bar \alpha }_t}} \bm{x}_0 + \sqrt {1 - {{\bar \alpha }_t}} \epsilon$.
Based on that, we can train a noise prediction function ${\epsilon _\theta }\left( {{\bm{x}_t},t} \right)$ to simulate the target distribution.

Though the target domain is unlabeled, we expect that labels of generated target samples can be controlled.
To make the DPM class-conditional, the pseudo label ${\bar y}^{td}$ assigned by the pretrained classifier in step 1 can be injected into the DPM in the same way as the time step $t$ through the label embedding $e_l=\mathrm{emb}({\bar y}^{td})$, where $\mathrm{emb}(\cdot)$ denotes the embedding function.
The entire embedding can be obtained by the sum of the time embedding 
and label embedding.
Then the embedding will be fed into residual blocks of the model ${{{\rm{\epsilon}}_\theta }\left( {\bm{x}_t^{td},{{\bar y}^{td}},t} \right)}$ to achieve class-conditional generation.

With the pretrained DPM on ImageNet \cite{nichol2021improved}, the model can converge faster in the target domain while reducing the required target domain samples.
To train CDPM, we use the gradient descent operation to minimize the following loss as
\begin{equation}
\begin{split}
\begin{array}{l}
\label{eqn4}
\mathcal{L}_d={\mathbb{E}_{{\epsilon \sim \mathcal{N}\left( {0,\bf{I}} \right)}}} {\left\| {\epsilon  - {\epsilon _\theta }\left( {{\bm{x}_t^{td}}, {{\bar y}^{td}}, t} \right)} \right\|^2},
\end{array}
\end{split}
\end{equation}
where $\theta$ denotes the parameters of the model $\epsilon _\theta$, $t$ is sampled from a uniform distribution $\mathrm{Uniform}\left( {\left\{ {1,...,T} \right\}} \right)$,
and ${\bm{x}_t^{td}} = \sqrt {{{\bar \alpha }_t}} \bm{x}_0^{td} + \sqrt {1 - {{\bar \alpha }_t}} \epsilon$.
When the loss $\mathcal{L}_d$ converges, we get an optimal CDPM $\epsilon _\theta^*$.

\subsection{Sampling and Training Final Classifier}
After the training of CDPM, we generate pseudo target samples with pseudo labels ${\mathcal{D}_g} = \{ ({\bm{x}}_k^g,\bar y_k^g)\} _{k = 1}^{N_g}$, where $N_g$ denotes the number of the generated target samples.
The distribution of $\mathcal{D}_g$ could be similar to that of the real target domain, thus alleviating the problem of scarce and unlabeled target samples.
Due to the slow sampling speed of DPMs, we adopt the DPM-Solver++ \cite{lu2022dpm} to greatly speed up the guided sampling process of CDPM, and so the sampling process is very efficient.
Specifically, given the initial noise $\bm{x}_T$ and time steps $\left\{ {{t_i}} \right\}_{i = 0}^M$,
the multi-step second-order solver for CDPM can be obtained as
\begin{align}
{W_i} \displaystyle &\!\leftarrow\!\! \left(\!{1\! +\!\frac{1}{{2{r_i}}}} \!\right)\!{\epsilon_\theta \!}\left( {{{\bm{x}}_{{t_{i - 1}}}}, l, {t_{i - 1}}} \right) \!-\! \frac{{{h_i}}}{{2{h_{i - 1}}}}{\epsilon_\theta }\!\left( {{{\bm{x}}_{{t_{i - 2}}}}, l, {t_{i - 2}}} \right)\nonumber\\
{{\bm{x}}_{{t_i}}} \displaystyle &\!\leftarrow\! \frac{{{\sigma _{{t_i}}}}}{{{\sigma _{{t_{i - 1}}}}}}{{\bm{x}}_{{t_{i - 1}}}} - {\alpha _{{t_i}}}\left( {{e^{ - {h_i}}} - 1} \right){W_i},\label{eqn2}
\end{align}
where ${h_i} = {\lambda _{{t_i}}} - {\lambda _{{t_{i - 1}}}}$ for $i=1,...,M$, ${\lambda _t} = \log \left( {\frac{{{\alpha _t}}}{{{\sigma _t}}}} \right)$, $M$ denotes the required time steps, and $l$ denotes the category we want to generate.

Considering that noisy pseudo labels may be contained in the generated target domain $\mathcal{D}_g$, we combine the original source domain $\mathcal{D}_s$ and $\mathcal{D}_g$ together as the augmented source domain ${\mathcal{D}_{\hat{s}}}$, and use some existing UDA method to transfer from the augmented source domain ${\mathcal{D}_{\hat{s}}}$ to the target domain $\mathcal{D}_t$ by solving the following objective function as
\begin{align}
\label{eqn3}
\!\!\!\! {g^*}\!\!\!=\! \displaystyle \mathop {\arg \min \! }\limits_{g} \frac{1}{{{N_{\hat{s}}}}}\!\!\sum\limits_{i = 1}^{{N_{\hat{s}}}} \!\mathcal{L}_c\! \left( {g\!\left( {{\bm{x}_i^{{\hat{s}d}}}} \right)\!,\!{y_i^{\hat{s}d}}} \right)
 \!\!+\!\!\lambda {{R(T(\mathcal{D}_{\hat{s}}),\!T(\mathcal{D}_t)\!)}},\!\!\!\!
\end{align}
where $N_{\hat{s}}$ denotes the number of samples in the augmented source domain.
After optimizing problem (\ref{eqn3}), the final target classifier $g^*$ can be obtained for the target domain.

\subsection{Discussions}
Traditional discrepancy-based and adversarial-based UDA methods focus on learning domain-invariant feature representations based on the source and target data.
However, the target data is limited and cannot represent the real target distribution.
The transfer performance of UDA models is usually limited by target samples.
Most of the existing generation-based UDA methods are based on image-to-image translation across domains \cite{hoffman2018cycada, yang2020bi} or generating data in an intermediate domain  \cite{cui2020gradually, na2021fixbi, gao2022middlegan}.
Those methods need to design complex mechanisms to maintain the detailed and semantic information of generated images.
Besides, the aforementioned GAN-based methods are unstable in adversarial training and require careful hyperparameter and regularizer selection to achieve the convergence \cite{dhariwal2021diffusion, murphy2023probabilistic}.


Different from those GAN-based methods, the proposed diffusion-based DTS framework directly generates pseudo target samples that could obey the target distribution without adversarial training.
\cite{nichol2021improved} has shown that DPMs are better at covering the modes of a distribution than GANs, which well meets the needs of target data generation here.
And the category and the number of samples generated can be flexibly controlled.
Moreover, the easy-to-implement plug-and-play DTS framework can be embedded into any UDA method to improve the transfer performance as shown in our experiments.

Next, we discuss the validity of the proposed DTS framework through the UDA theory \cite{ben2010theory}.
	
\emph{Proposition 1 \cite{ben2010theory}:} For a hypothesis space $\mathcal{H}$, we have
\begin{equation} \label{eqn-thrm}
r_t\left( h \right)\leq  r_s\left( h \right)+\frac{1}{2}{d_{\mathcal{H} \Delta\mathcal{H}} }\left( {\mathcal{S} ,\mathcal{T}} \right)+C,
\end{equation}
where $r_s\left( h \right)$ and $r_t\left( h \right)$ denote the expected risks in the source and target domain, respectively, ${d_{\mathcal{H} \Delta\mathcal{H}} }\left( {\mathcal{S},\mathcal{T}} \right)$ denotes the $\mathcal{H} \Delta \mathcal{H}$-distance between the source and target distributions $\mathcal{S}$ and $\mathcal{T}$, and
$C$ denotes a negligible term \cite{ganin2015unsupervised, long2015learning}.
In DTS, with generated target samples, we can have
\begin{small}
\begin{align}
\label{eqn-thrm1}
{r_t}(\hat{h}) \leq& {r_{\hat{s}}}( {\hat h} ) + \frac{1}{2}{d_{{\rm{{\cal H}}}\Delta {\rm{{\cal H}}}}}( {\hat {\rm{{\cal S}}},{\rm{{\cal T}}}} ) + C \nonumber \\
=& {r_{\hat s}}( {\hat h} ) + {r_{\hat s}}\left( h \right) - {r_{\hat s}}\left( h \right) + \frac{1}{2}{d_{{\rm{{\cal H}}}\Delta {\rm{{\cal H}}}}}( {\hat {\rm{{\cal S}}},{\rm{{\cal T}}}} ) + C \nonumber \\
\leq & {r_{\hat s}}\left( h \right) + \frac{1}{2}{d_{{\rm{{\cal H}}}\Delta {\rm{{\cal H}}}}}( {\hat {\rm{{\cal S}}},{\rm{{\cal T}}}} ) + C \nonumber \\
=& {r_s}\left( h \right) + {r_g}\left( h \right) + \frac{1}{2}{d_{{\rm{{\cal H}}}\Delta {\rm{{\cal H}}}}}( {\hat {\rm{{\cal S}}},{\rm{{\cal T}}}} ) + C \nonumber \\ \nonumber
\leq& {r_s}\left( h \right) + {r_g}\left( h \right) + \frac{1}{2}\big[ \alpha {d_{{\rm{{\cal H}}}\Delta {\rm{{\cal H}}}}}( {{\rm{{\cal S}}},{\rm{{\cal T}}}} ) \\ \nonumber
 & + \left( {1 - \alpha } \right){d_{{\rm{{\cal H}}}\Delta {\rm{{\cal H}}}}}( {{{\rm{{\cal S}}}_g},{\rm{{\cal T}}}} ) \big] + C\\ \nonumber
=& {r_s}\left( h \right) + \frac{1}{2}{d_{{\rm{{\cal H}}}\Delta {\rm{{\cal H}}}}}\left( {{\rm{{\cal S}}},{\rm{{\cal T}}}} \right)  + C +\! {r_g\!}\left( h \right)\\
& + \frac{1}{2}(1 - \alpha )\left[ {{d_{{\rm{{\cal H}}}\Delta {\rm{{\cal H}}}}}\left( {{{\rm{{\cal S}}}_g},{\rm{{\cal T}}}} \right) \!-\! {d_{{\rm{{\cal H}}}\Delta {\rm{{\cal H}}}}}\left( {{\rm{{\cal S}}},{\rm{{\cal T}}}} \right)} \right],
\end{align}
\vspace{-0.3cm}
\end{small}

\noindent
where $\hat {\rm{{\cal S}}}$ and ${{\rm{{\cal S}}}_g}$ denote the distribution of the augmented source domain ${\mathcal{D}_{\hat{s}}}$ and the generated domain $\mathcal{D}_g$, respectively,
${r_g}(h)$ denotes the expected risk in the generated domain $\mathcal{D}_g$,
$\alpha$ is a constant in $[0,1]$, and
$\hat{h} = \mathop{\arg\min}\limits_{h \in \mathcal{H}} r_{\hat{s}}(h)$ represents the classifier trained in the augmented source domain ${\mathcal{D}_{\hat{s}}}$.
Note that ${r_s}\left( h \right) + \frac{1}{2}{d_{{\rm{{\cal H}}}\Delta {\rm{{\cal H}}}}}\left( {{\rm{{\cal S}}},{\rm{{\cal T}}}} \right)+C$ is the same as the right-hand side of \emph{Proposition 1}.
As $\mathcal{S}_g$ is generated to approximate $\mathcal{T}$, it is expected that $\mathcal{S}_g$ has a lower discrepancy to $\mathcal{T}$ than $\mathcal{S}$,
i.e., $ {{d_{{\rm{{\cal H}}}\Delta {\rm{{\cal H}}}}}\left( {{{\rm{{\cal S}}}_g},{\rm{{\cal T}}}} \right) < {d_{{\rm{{\cal H}}}\Delta {\rm{{\cal H}}}}}\left( {{\rm{{\cal S}}},{\rm{{\cal T}}}} \right)}$.
When ${r_g}\left( h \right)$ is small, ${r_t}({\hat h})$ will have a lower upper-bound when compared with $r_t (h)$, which could explain the superior performance of the DTS method from the theoretical perspective.

\section{Experiments}

In this section, we empirically evaluate the proposed DTS framework.
\subsection{Experimental Settings}

\noindent\textbf{Datasets.} Experiments are conducted on three benchmark datasets, including Office-31 \cite{saenko2010adapting}, Office-Home \cite{venkateswara2017deep}, and VisDA-2017 \cite{peng2017visda}.
The \textbf{Office-31} dataset contains 4,110 images in 31 categories of three distinct domains: Amazon (A), DSLR (D), and Webcam (W).
From these three domains, we build six different transfer tasks, i.e., A$\rightarrow$W, D$\rightarrow$W, W$\rightarrow$D,  A$\rightarrow$D, D$\rightarrow$A, W$\rightarrow$A.
The \textbf{Office-Home} dataset contains 15,500 images in total from 65 categories of four image domains: Art (Ar), Clipart (Cl), Product (Pr), and Real-World (Rw). On this dataset, we build 12 transfer tasks in the four domains.
The \textbf{VisDA-2017} dataset is a large-scale synthetic-to-real  benchmark dataset for UDA with 12 categories.
It contains more than 150k images in the source domain and 50k images in the target domain.

\noindent\textbf{Implementation details.}
We use the pretrained CDPM on ImageNet provided by \cite{nichol2021improved}. $\epsilon _\theta$ adopts a U-Net model \cite{ronneberger2015u}.
The number of target samples generated for each category in DTS is set to 200, 200, and 2000 on the Office-31, Office-Home, and VisDA datasets, respectively.
The resolution of generated target samples is 256$\times$256.
The DPM-Solver++ is used to ensure stable class-conditional sampling.
For a fair comparison, the same backbone is used for all the methods on one dataset.
Specifically, ResNet-50 \cite{he2016deep} is used on the Office-31 and Office-Home datasets, and ResNet-101 is used on the VisDA-2017 dataset.
We use mini-batch stochastic gradient descent (SGD) optimizer with a momentum of 0.9 and the same learning rate annealing strategy as \cite{ganin2016domain}.
All experiments are implemented on a NVIDIA V100 GPU.

\noindent{\bf Baselines.} We compare with state-of-the-art UDA methods, including
DANN \cite{ganin2016domain}, AFN \cite{AFN}, CDAN \cite{long2018conditional}, MDD \cite{MDD}, SDAT \cite{rangwani2022closer}, MCC \cite{jin2020minimum}, and ELS \cite{zhang2023free}.
In particular, we combine the proposed DTS with MCC and ELS to evaluate its effectiveness.
We also compare with ERM \cite{vapnik1999nature}, which trains a classifier on the source domain and applies it directly to the target domain without domain alignment.

\subsection{Experimental Results}
In this section, we analyze results on the three benchmark datasets.
The DTS framework is easily embedded into state-of-the-art UDA models such as MCC \cite{jin2020minimum} and ELS \cite{zhang2023free}, i.e., MCC+DTS and ELS+DTS.

\noindent\textbf{Office-31.}
According to results shown in Table \ref{office31} for the Office-31 dataset, we can see that the proposed DTS method outperforms baseline methods by a large margin.
Specifically, MCC+DTS achieves an average accuracy of 90.6\%, and the proposed DTS framework brings a  performance improvement of 0.8\% compared with the original MCC method.
ELS+DTS achieves an average accuracy of 91.0\%, which outperforms the original ELS method by 0.6\%.
Moreover, for the most difficult transfer task W$\rightarrow$A, where the difficulty of a transfer task is measured via the average classification accuracy of all the models in comparison, the proposed DTS framework achieves performance improvements of 2.5\% and 2.1\% over MCC and ELS, respectively.
Therefore, DTS can help improve the performance of UDA methods.

\begin{table}[!t]\small
\centering
\caption{Accuracy comparison (\%) on the Office-31 dataset with ResNet-50 as the backbone. $\uparrow$ denotes the accuracy improvement brought by the DTS framework over the corresponding baseline (i.e., MCC or ELS). The best performance of each task is marked in bold.}
\label{office31}
\setlength{\tabcolsep}{1.5mm}{
\resizebox{\columnwidth}{!}{
\begin{tabular}{ccccccc @{\hskip 0.2in} c}
\toprule Method & A$\rightarrow$W        & D$\rightarrow$W        & W$\rightarrow$D        & A$\rightarrow$D        & D$\rightarrow$A        & W$\rightarrow$A        & Average \\
\midrule
ERM \cite{vapnik1999nature}  & 77.0 & 96.6 & 99.2 & 82.8 & 64.1 & 64.1 & 80.3 \\
DANN \cite{ganin2016domain} & 89.3 & 98.0 & \textbf{100.0} & 83.5 & 74.0 & 74.3 & 86.5 \\
AFN \cite{AFN}  & 91.3 & 98.7 & \textbf{100.0} & 95.6 & 72.1 & 70.7 & 88.1 \\
CDAN \cite{long2018conditional} & 92.6 & 98.5 & \textbf{100.0}  & 92.2 & 75.7 & 73.1 & 88.7 \\
MDD \cite{MDD}  & 94.6 & 98.5 & \textbf{100.0}  & 94.0 & 75.7 & 73.9 & 89.4 \\
SDAT \cite{rangwani2022closer} & 88.8 & \textbf{99.0} & \textbf{100.0} & 93.8 & 76.3 & 72.8 & 88.4 \\
\midrule
MCC \cite{jin2020minimum}  & 93.7 & 98.4 & 99.6  & 95.2 & 76.3 & 75.5 & 89.8 \\
MCC+DTS  & \textbf{94.7} & 98.4 & 99.6  & 95.4 & 77.8 & \textbf{78.0} & 90.6 \\
$\uparrow $  & 1.0 & 0.0 & 0.0 & 0.2 & 1.5 & 2.5 & 0.8 \\
\midrule
ELS \cite{zhang2023free}  & 94.3 & 98.9 & \textbf{100.0} & 95.6 & 78.5 & 75.0 & 90.4 \\
ELS+DTS      & 94.5 & \textbf{99.0} & \textbf{100.0} & \textbf{96.0} & \textbf{79.3} & 77.1 & \textbf{91.0} \\
$\uparrow $  & 0.2 & 0.1 & 0.0 & 0.4 & 0.8 & 2.1 & 0.6 \\
\bottomrule
\end{tabular}}}
\end{table}

\begin{table*}[!tbph]\small
\centering
\caption{Accuracy comparison (\%) on the Office-Home dataset with ResNet-50 as the backbone. $\uparrow$ denotes the accuracy improvement brought by the DTS framework  over the corresponding baseline (i.e., ELS). The best performance of each task is marked in bold.}
\label{officehome}
\setlength{\tabcolsep}{1mm}{
\begin{tabular}{ccccccccccccc @{\hskip 0.05in} c}
\toprule Method & Ar$\rightarrow$Cl   & Ar$\rightarrow$Pr  & Ar$\rightarrow$Rw & Cl$\rightarrow$Ar & Cl$\rightarrow$Pr & Cl$\rightarrow$Rw & Pr$\rightarrow$Ar & Pr$\rightarrow$Cl & Pr$\rightarrow$Rw & Rw$\rightarrow$Ar & Rw$\rightarrow$Cl & Rw$\rightarrow$Pr & Average \\
\midrule
ERM \cite{vapnik1999nature}  & 44.2 & 67.2 & 74.6 & 53.3 & 61.8 & 64.4 & 51.9 & 38.8 & 73.0 & 64.3 & 43.6 & 75.4 & 59.4  \\
DANN \cite{ganin2016domain} & 52.1 & 63.3 & 73.5 & 56.9 & 67.4 & 67.8 & 57.8 & 54.8 & 78.7 & 71.4 & 60.6 & 80.0 & 65.4   \\
AFN \cite{AFN} & 52.6 & 72.6 & 76.9 & 64.9 & 71.5 & 73.0 & 63.7 & 51.5 & 77.8 & 72.1 & 57.7 & 82.3 & 68.0 \\
CDAN \cite{long2018conditional} & 53.9 & 72.4 & 78.6 & 61.9 & 70.8 & 72.3 & 63.5 & 55.6 & 80.9 & 75.0 & 61.5 & 83.2 & 69.1 \\
MDD \cite{MDD} & 55.7 & 75.9 & 79.1 & 63.0 & 73.7 & 74.1 & 63.0 & 55.6 & 79.8 & 73.8 & 61.4 & 84.0 & 69.9 \\
SDAT \cite{rangwani2022closer} & 58.4 & 77.9 & 81.6 & 66.2 & 76.8 & 76.4 & 62.2 & 56.2 & 82.6 & 75.7 & 62.0 & 85.3 & 71.8 \\
\midrule
MCC  & 56.2 & 79.5 & 82.5 & 68.1 & 76.5 & 78.0 & 66.9 & 54.9 & 82.1 & 73.6 & 61.9 & 85.5 & 72.1   \\
MCC+DTS      & 56.7 & \textbf{80.8} & \textbf{83.8} & \textbf{69.6} & 79.5 & \textbf{81.4} & \textbf{67.7} & 57.1 & \textbf{83.9} & 73.6 & 62.1 & \textbf{86.4} & 73.5\\
$\uparrow $      & 0.4 & 1.3 & 1.3 & 1.5 & 3.0 & 3.4 & 0.8 & 2.2 & 1.8 & 0.0  & 0.3 & 0.9 & 1.4   \\
\midrule
ELS \cite{zhang2023free}  & 57.2 & 77.2 & 82.0 & 66.5 & 77.2 & 76.7 & 62.3 & 56.3& 82.2 &  75.6 & 63.9 & 85.4 & 71.9  \\
ELS+DTS      & \textbf{59.9} & 78.8 & 83.2 & 67.8 & \textbf{81.1} & 80.6 & 63.9 & \textbf{57.9} & 83.2 & \textbf{75.9} & \textbf{64.2} & 86.1 & \textbf{73.6}\\
$\uparrow $      & 2.7 & 1.6 & 1.2 & 1.3 & 3.9 & 3.9 & 1.6 & 1.6 & 1.0 & 0.3 & 0.3 & 0.7 & 1.7 \\
\bottomrule
\end{tabular}}
\end{table*}

\begin{table*}[!tbph]\small
\centering
\caption{Accuracy comparison (\%) on the VisDA-2017 dataset with ResNet-101 as the backbone. $\uparrow$ denotes the accuracy improvement brought by the DTS framework  over the corresponding baseline (i.e., MCC or ELS). The best performance of each task is marked in bold.}
\label{visda}
\setlength{\tabcolsep}{1.5mm}{
\begin{tabular}{ccccccccccccc @{\hskip 0.2in} c}
\toprule Method & aero & bicycle & bus & car & horse & knife & motor & person & plant & skate & train & truck & mean \\
\midrule
ERM \cite{vapnik1999nature}  & 86.9 & 20.8 & 58.4 & 74.3 & 76.3 & 15.4 & 85.8 & 17.5 & 82.5 & 29.2 & 80.2 & 5.5	& 52.7  \\
DANN \cite{ganin2016domain} & 94.3 & 76.1 & 83.6 & 44.4 & 86.3 & 92.8 & 87.4 & 78.4 & 89.0 & 88.3 & 87.5 & 46.1 & 79.5   \\
AFN \cite{AFN} & 92.2 & 54.8 & 82.8 & 72.2 & 90.8 & 74.4 & 92.3 & 70.9 & \textbf{94.7} & 56.4 & \textbf{89.1} & 25.3	& 74.6 \\
CDAN \cite{long2018conditional} & 95.0 & 75.9 & \textbf{85.7} & 57.0 & 90.9 & 95.9 & 89.9 & 75.6 & 85.8 & 89.5 & 88.5 & 41.1	&80.9 \\
MDD \cite{MDD} & 89.3 & 70.2 & 82.3 & 66.7 & 92.6 & 94.4 & \textbf{93.3} & 78.9 & 92.6 & 88.0 & 82.3 & 49.0	& 81.6 \\
SDAT \cite{rangwani2022closer} & 94.7 & 85.1 & 73.2 & 67.0 & 93.0 & 94.7 & 89.2 & 80.7 & 90.8 & 93.0 &  84.1 & 56.0 & 83.5  \\
\midrule
MCC \cite{jin2020minimum}  & 95.1 & 84.9 & 73.0  & 69.9 & 93.2 & 95.6 & 87.6 & 79.0 & 90.4 & 89.9 & 84.9 & 55.7 & 83.3 \\
MCC+DTS   & 96.4  & \textbf{87.2} & 82.2 & \textbf{78.2} & 94.6 & \textbf{96.8} & 88.7 & \textbf{82.6} & 92.4 & \textbf{93.8} & 86.6 & 57.8 & \textbf{86.4} \\
$\uparrow $  & 1.3 & 2.3 & 9.2 & 8.3 & 1.4 & 1.2 & 1.1 & 3.6 & 2.0 & 3.9 & 1.7 & 2.1 & 3.1 \\
\midrule
ELS \cite{zhang2023free} & 95.3 & 84.9 & 75.2 & 66.3 & 93.0 & 93.8 & 88.4 & 79.4 & 90.4 & 92.5 & 83.8 & 57.7 & 83.4 \\
ELS+DTS   & \textbf{96.6}  & 86.1 & 82.2 & 68.8 & \textbf{95.4} & 96.3 & 90.8 & 82.5 & 92.7 & 93.3 &87.9 & \textbf{60.5} & 86.1 \\
$\uparrow $  & 1.3 & 1.2 & 7.0 & 2.5 & 2.4 & 2.5 & 2.4 & 3.1 & 2.3 & 0.8 & 4.1 & 2.8 & 2.7 \\
\bottomrule
\end{tabular}}
\end{table*}

\noindent\textbf{Office-Home.} According to results shown in Table \ref{officehome} for the Office-Home dataset,
we can see that the proposed DTS framework brings performance improvement for most transfer tasks, which demonstrates its effectiveness.
Specifically, ELS+DTS achieves an average accuracy of 73.6\%, which improves the average accuracy of ELS by 1.7\%.
Especially, in transfer tasks Cl$\rightarrow$Pr and Cl$\rightarrow$Rw, DTS brings significant performance improvements (i.e., 3.9\%) over ELS.
Moreover, MCC achieves an average accuracy improvement of 1.4\% over the original MCC method, which further demonstrates the effectiveness of the DTS method.
Overall, the proposed DTS framework has shown its effectiveness on this dataset.

\noindent\textbf{VisDA-2017.} The results on the VisDA-2017 dataset are shown in Table \ref{visda}.
It can be seen that MCC+DTS achieves an average classification accuracy of 86.4\%, which outperforms the original MCC by a large margin of 3.1\%.
ELS+DTS achieves an average classification accuracy of 86.1\%, which outperforms the origin ELS by 2.7\%. Furthermore, the performance of all categories has been improved.
In particular, DTS obtains a significant improvement of 9.2\% on the `bus' category when compared with MCC.
Moreover, even the smallest accuracy improvement of MCC+DTS over MCC for each class is 1.1\%, which further demonstrates the superiority of the proposed DTS framework.

\begin{figure*}[t]
\centering
\includegraphics[width=\textwidth]{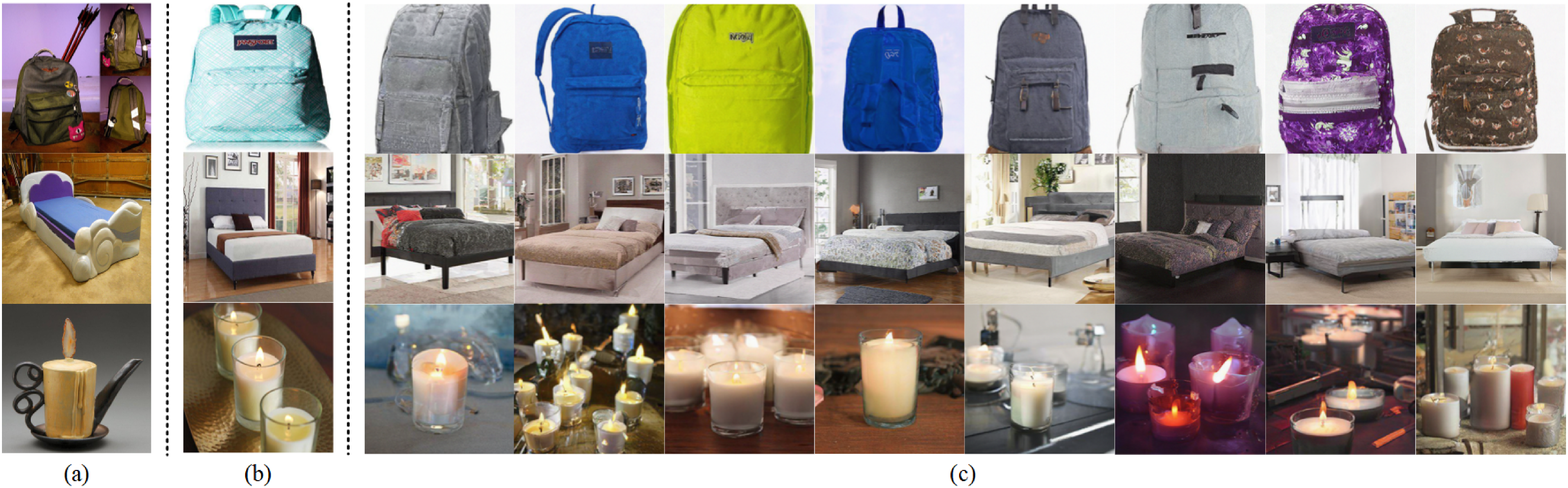}
\caption{Real and generated images for the transfer task Ar$\rightarrow$Cl on the Office-Home dataset. (a) and (b) show the real source and target images, respectively, and (c) shows the generated target samples.}
\label{generate1}
\end{figure*}

\begin{figure*}[t]
\centering
\includegraphics[width=\textwidth]{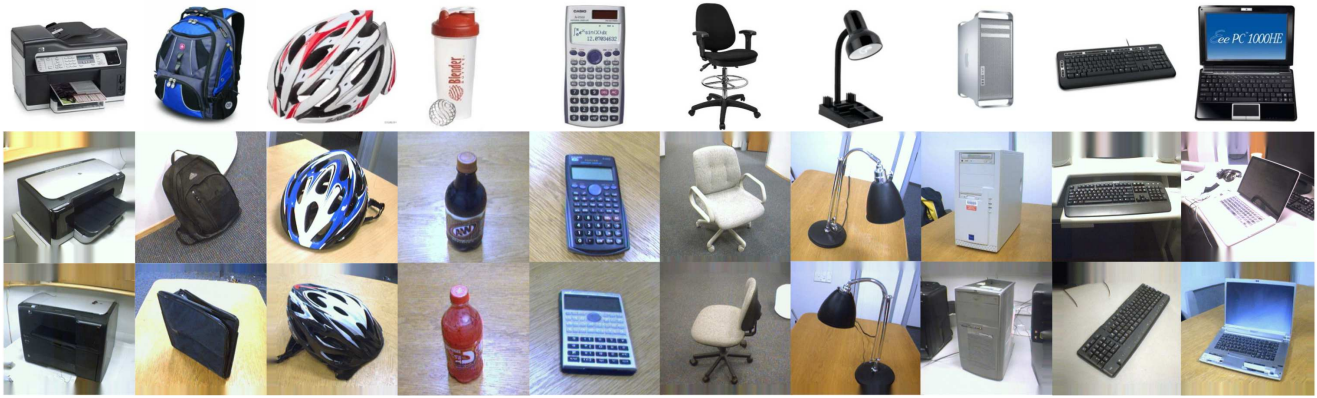}
\caption{Real and generated images for the transfer task A$\rightarrow$W on the Office-31 dataset. The first row shows the source samples. The second row shows the target samples.  The third row shows the generated target samples.}
\label{generate}
\end{figure*}

\begin{figure*}[t]
\centering
\includegraphics[width=\textwidth]{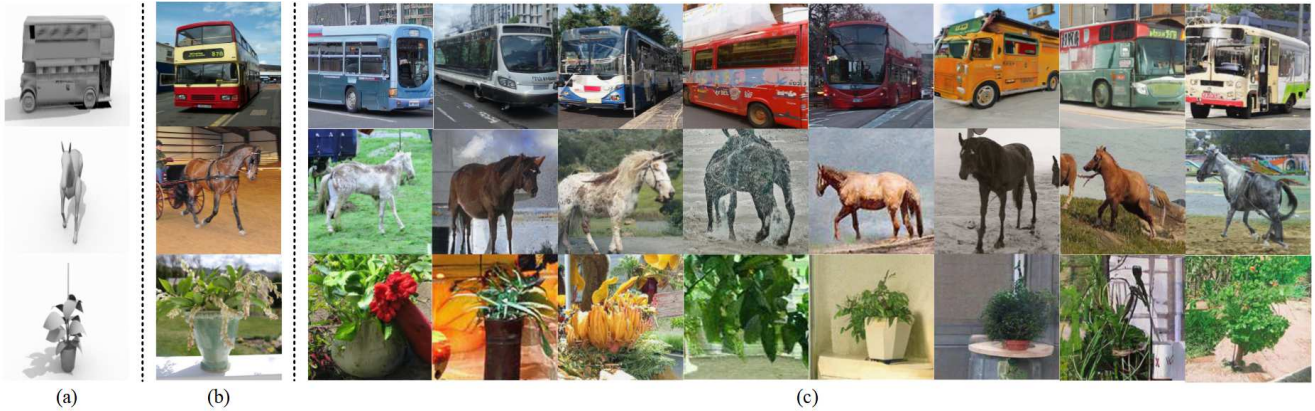}
\caption{Real and generated images on the VisDA dataset. (a) and (b) show the real source and target images, respectively, and (c) shows the generated target samples.}
\label{visda_generate}
\end{figure*}

\begin{table*}[!t]\small
\centering
\caption{Ablation studies (in terms of the classification accuracy (\%)) on the Office-Home dataset for UDA with ResNet-50 as the backbone. ELS is the baseline method. The last line is our method. The best performance of each task is marked in bold.}
\label{ablation}
\setlength{\tabcolsep}{1mm}{
\resizebox{\textwidth}{!}{
\begin{tabular}{ccccccccccccc @{\hskip 0.05in} c}
\toprule Ablation of ELS+DTS & Ar$\rightarrow$Cl   & Ar$\rightarrow$Pr  & Ar$\rightarrow$Rw & Cl$\rightarrow$Ar & Cl$\rightarrow$Pr & Cl$\rightarrow$Rw & Pr$\rightarrow$Ar & Pr$\rightarrow$Cl & Pr$\rightarrow$Rw & Rw$\rightarrow$Ar & Rw$\rightarrow$Cl & Rw$\rightarrow$Pr & Average \\
\midrule
ELS \cite{zhang2023free}  & 57.2 & 77.2 & 82.0 & 66.5 & 77.2 & 76.7 & 62.3 & 56.3& 82.2 &  75.6 & 63.9 & 85.4 & 71.9  \\
ELS+DTS (w/o generation) & 58.6 & 77.6 & 81.2  & 66.2 & 76.4 & 76.0 & 61.9 & 56.2 & 82.6 & 75.2 & 62.7 & 84.9 & 71.6 \\
ELS+DTS (w/o original source domain) &57.5 &77.5	&82.4 &66.8&80.1	&79.6	&62.4	&54.2	&81.6	& 72.9 & 60.8	&84.8	&71.7 \\									
ELS+DTS (Finetune the pretrained classifier) & 59.1 & 78.0 & 83.0  &\textbf{68.1} &79.8 & 79.3 & 62.7 & 57.5 & 82.8 & 75.8 & \textbf{64.2} & \textbf{86.4} & 73.1 \\
ELS+DTS (Train from scratch)    & \textbf{59.9} & \textbf{78.8} & \textbf{83.2} & 67.8 & \textbf{81.1} & \textbf{80.6} & \textbf{63.9} & \textbf{57.9} & \textbf{83.2} & \textbf{75.9} & \textbf{64.2} & 86.1 & \textbf{73.6}\\
\bottomrule
\end{tabular}}}
\end{table*}

\subsection{Ablation Studies}
To analyze the effectiveness of the proposed DTS method in detail, ablation experiments are conducted on the Office-Home dataset.
`ELS+DTS (w/o generation)' means that the target samples with pseudo labels given by the classifier trained by ELS and the original source domain are combined to train the final classifier without using the generative model.
`ELS+DTS (w/o original source domain)' means that  generated target samples with pseudo labels are treated as a new source domain and then transferred  to the target domain via the ELS directly without the original source domain.
Different from the proposed DTS framework that trains a UDA model from scratch to transfer the augmented source domain to the target domain, `ELS+DTS (Finetune the pretrained classifier)' means that the pretrained classifier in step 1 is used for parameter initialization.




As shown in Table \ref{ablation}, we can see that if we directly transfer from generated target samples to the target domain as ELS+DTS (w/o original source domain) did, a small degree of negative transfer (i.e., 0.2\% in terms of average classification accuracy) still occurs. This happens due to noisy pseudo labels of generated target samples.
Hence, the proposed DTS framework combines the original source domain and the generated pseudo target domain as the augmented source domain and transfers it to the target domain.
On one hand, correct source information could help alleviate label noises of generated samples.
On the other hand, generated target samples can inject more useful information about the target domain into the source domain to reduce the domain discrepancy and improve the transfer effects.
ELS+DTS (w/o generation) performs worse on average than ELS+DTS (Train from scratch), which further demonstrates that generated target samples contain useful information of the target domain.

In the proposed DTS framework, we train a UDA model from scratch in step 3 to transfer from the augmented source domain to the target domain without using the pretrained classifier in step 1.
According to Table \ref{ablation}, we can see that training from scratch performs slightly better with an average improvement of 0.5\% than finetuning the pretrained classifier.
As a result, finetuning the pretrained classifier can converge faster and reduce the training time but with slightly lower transfer performance on average.
Hence, to trade off between the accuracy and time complexity, we can choose to finetune the pretrained classifier or train the UDA model from scratch according to the actual situation.

\subsection{Analyses on Generated Results}
To demonstrate the fidelity and diversity of generated samples, we show in Figure~\ref{generate1} generative results of the transfer task Ar$\rightarrow$Cl on the Office-Home dataset.
It can be seen that the style of generated samples is similar to real target samples.
Moreover, the generated samples have high fidelity and diversity, so they can well simulate the distribution of the target domain.
Figure~\ref{generate} shows the generated target samples in the transfer task A$\rightarrow$W on the Office-31 dataset.
We can see that generated samples are of high quality and similar to real target samples, and that the categories of generated images can be well controlled. 
As the size of the Office-31 dataset is the smallest among three datasets, high fidelity generated target samples demonstrate the effectiveness of the proposed DTS method.
According to generative target samples on the VisDA dataset as shown in Figure~\ref{visda_generate},
we can see that the proposed DTS method can well simulate the distribution of the target domain while maintaining the diversity of generated target samples.

\begin{figure}[t]
\centering
\subfigure[A$\rightarrow$W]{
    \includegraphics[width=3.7cm]{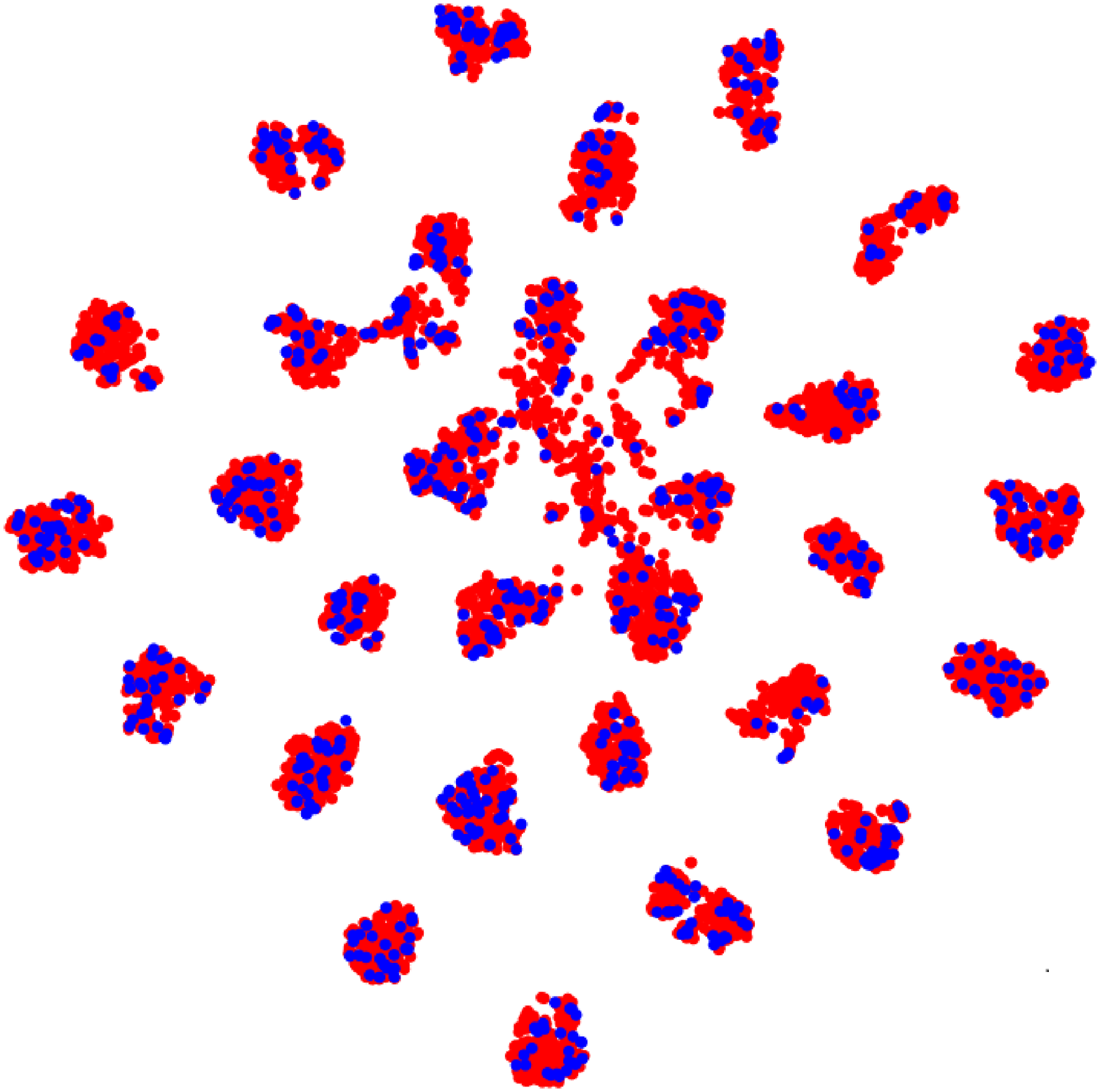}
    \label{x1}
}
\quad
\subfigure[D$\rightarrow$W]{
    \includegraphics[width=3.7cm]{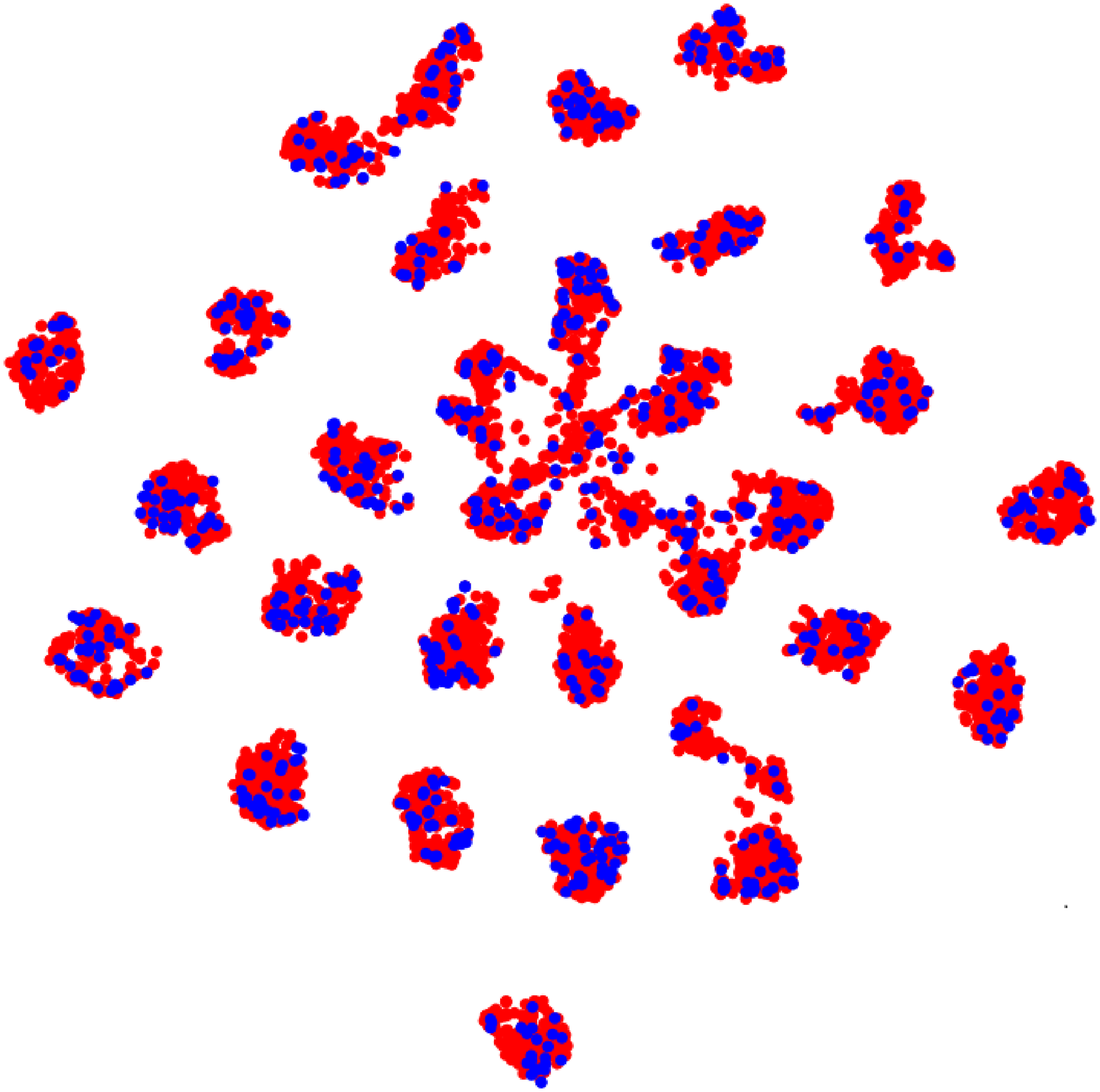}
    \label{x2}
}
    \caption{Feature visualization via t-SNE on the Office-31 dataset. Red and Blue points denote the generated target samples and original target samples, respectively. Best viewed in color.}
\label{tSNE}
\end{figure}

Figure~\ref{tSNE} visualizes feature representations in the fully-connected layer of the final classifier on two transfer tasks of the Office-31 dataset via the t-SNE method \cite{van2008visualizing}.
It can be observed that generated target samples are very close to real target samples and hence they can help model the data distribution of the target domain as real target samples are limited.
On the other hand, due to the incorporation of generated target samples which can well approximate the data distribution of the target domain, the augmented source domain could have a smaller domain discrepancy to the target domain than that between the original source and target domains, and this makes the transfer tasks easier.
Hence, the performance of existing UDA methods can be improved.

\begin{table}[!t]\small
\centering
\caption{$\mathcal{A}$-distance across domains over transfer tasks on the Office-31 dataset.}
\label{Adistance}
\setlength{\tabcolsep}{1.5mm}{
\begin{tabular}{c @{\hskip 0.3in} cccc}
\toprule $\mathcal{A}$-distance & A$\rightarrow$W  & A$\rightarrow$D   & D$\rightarrow$A   & W$\rightarrow$A  \\
\midrule
$\mathcal{D}_s$ and $\mathcal{D}_t$ & 1.92 & 1.87 & 1.84 & 1.89 \\
$\mathcal{D}_g$ and $\mathcal{D}_t$ & 1.61 & 1.64 & 1.57 & 1.55 \\
$\mathcal{D}_{\hat{s}}$ and $\mathcal{D}_t$ & 1.71 & 1.78 & 1.59 & 1.64 \\
\bottomrule
\end{tabular}}
\end{table}

To further demonstrate the effectiveness of the proposed DTS framework, we introduce the $\mathcal{A}$-distance to measure the distribution discrepancy on the Office-31 dataset.
The $\mathcal{A}$-distance is defined as ${d_{\rm{\mathcal A}}}(p,q) = 2\left( {1 - 2\nu} \right)$, where $\nu$ denotes the error of a linear domain discriminator to distinguish source and target samples.
According to the results shown in Table \ref{Adistance}, we can see that the distribution of generated target samples is closer to that of the target domain than the original source domain.
Similar observations hold between the augmented source domain and the target domain.
Therefore, it is easier for UDA methods to transfer from the augmented source domain to the target domain, which further verifies the effectiveness of the DTS framework.

\subsection{Sensitivity Analysis}
\label{sec:sen_analysis}

\begin{figure}[t]
\centering
\subfigure{
    \includegraphics[width=3.95cm]{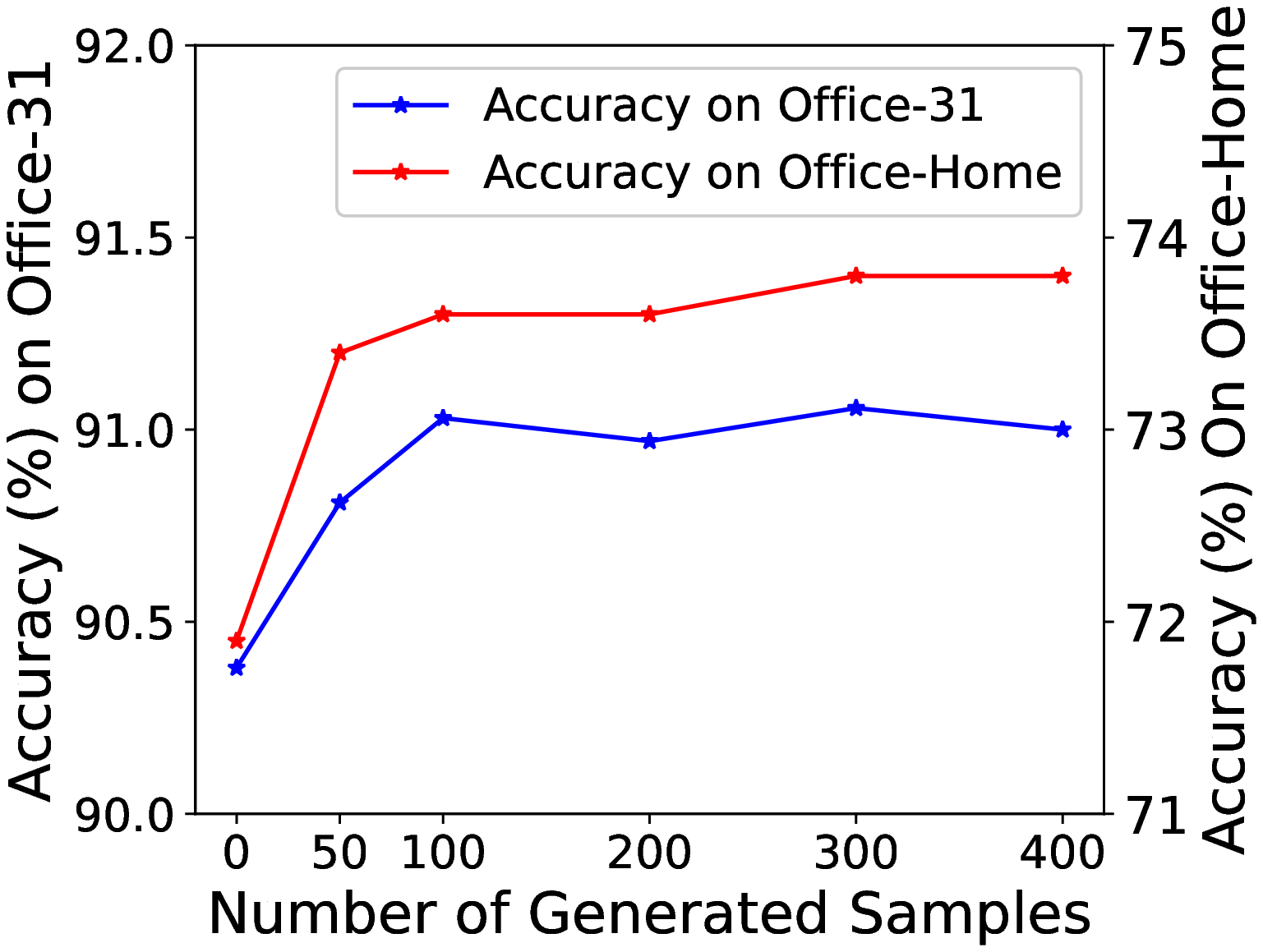}
    \label{x3}
}
\subfigure{
    \includegraphics[width=3.95cm]{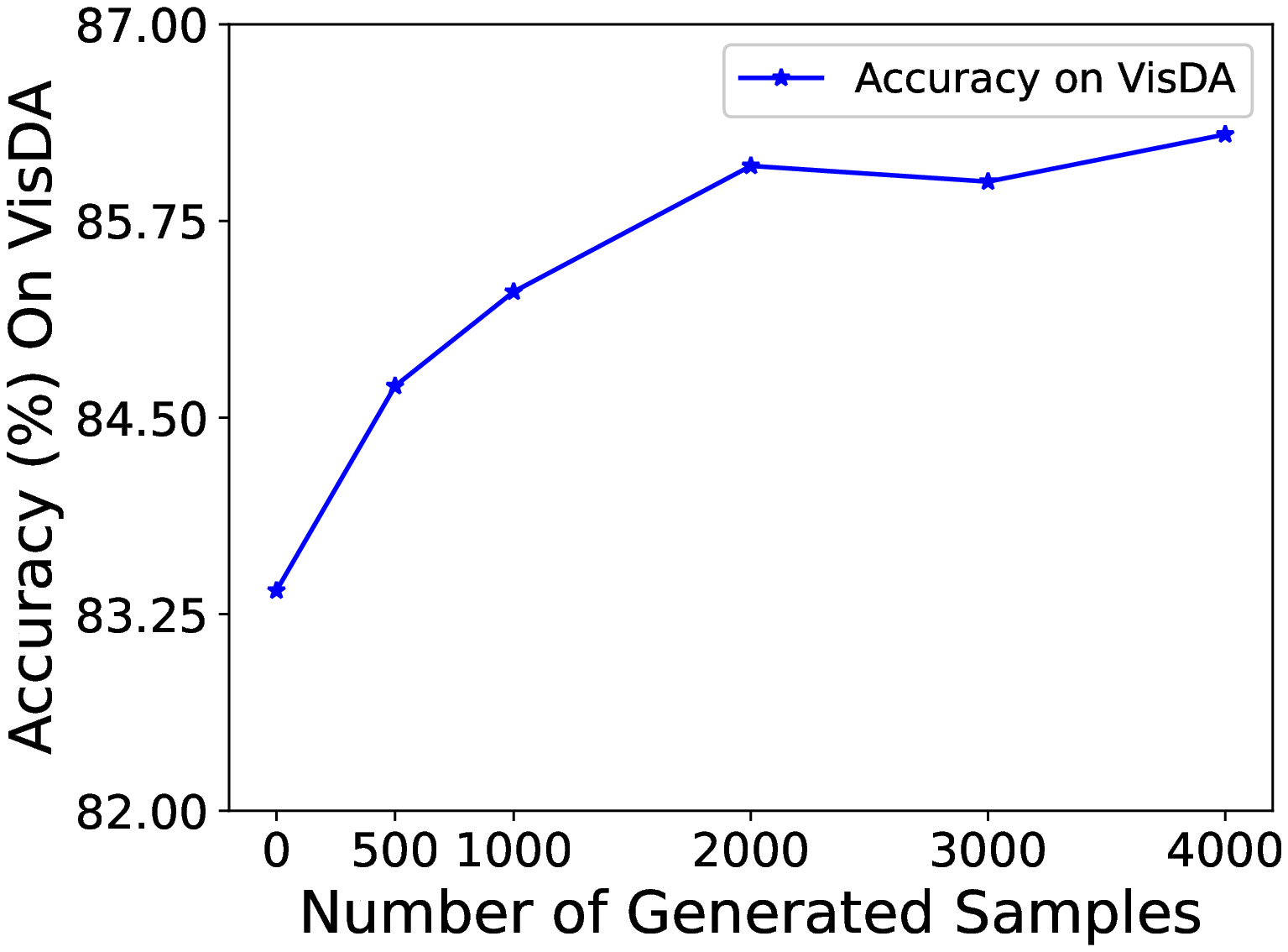}
    \label{x4}
}
    \caption{Sensitivity analysis on the number of generated target samples for each category on Office-31, Office-Home, and VisDA.}
\label{Sensitivity}
\end{figure}

In this section, we conduct sensitivity analysis to analyze the effect of the number of generated target samples for each category to the transfer performance on three datasets.
The average results of ELS+DTS are shown in Figure~\ref{Sensitivity}.
Note that when the number of generated target samples is 0, ELS+DTS degenerates to the original ELS.
According to results, we can see that increasing the number of generated samples improves the transfer performance, which proves the effectiveness of the proposed DTS method.
On the three datasets, we find that when the number of generated samples increases to a certain extent, 
the transfer performance tends to stabilize.
Therefore, we need to select an appropriate number of generated target samples to balance the transfer performance and the time consumption of generating target samples.
In this paper, the number of generated target samples for each category is set to 200, 200, and 2000 by default on the Office-31, Office-Home, and VisDA datasets, respectively, to achieve a good balance.

Detailed results for each transfer task in the sensitivity analysis are shown in Table \ref{office31_number}, \ref{officehome_number}, and \ref{visda_number}.
According to results, we can see that increasing the number of generated target samples improves the transfer performance, which demonstrates the effectiveness of the proposed DTS method.
For small-scale datasets such as Office-31 and Office-Home, only a few samples (e.g., 50-400 per category) need to be generated to improve the transfer performance, while for large-scale datasets such as VisDA, more samples (e.g., 2,000-4,000 per category) need to be generated to achieve stable performance improvement.

\begin{table}[!tbph]\small
\centering
\caption{Accuracy (\%) when varying the number of generated samples per category on the Office-31 dataset with ResNet-50 as the backbone.}
\label{office31_number}
\setlength{\tabcolsep}{1.5mm}{
\resizebox{\columnwidth}{!}{
\begin{tabular}{ccccccc @{\hskip 0.2in} c}
\toprule Method & A$\rightarrow$W        & D$\rightarrow$W        & W$\rightarrow$D        & A$\rightarrow$D        & D$\rightarrow$A        & W$\rightarrow$A        & Average \\
\midrule
0  & 94.3 & 98.9 & 100.0 & 95.6 & 78.5 & 75.0 & 90.4 \\
50  & 94.2 & 99.0 & 100.0 & 96.2 & 78.8 & 76.6 & 90.8  \\
100  & 94.3 & 99.0 & 100.0 & 96.0 & 79.6 & 77.3 & 91.0  \\
200  & 94.5 & 99.0 & 100.0 & 96.0 & 79.3 & 77.1 & 91.0  \\
300  & 94.7 & 98.9 & 100.0 & 96.0 & 79.7 & 77.1 & 91.1 \\
400  & 94.5 & 98.9 & 100.0 & 96.0 & 79.9 & 76.8 & 91.0  \\
\bottomrule
\end{tabular}}}
\end{table}

\begin{table*}[!tbph]\small
\centering
\caption{Accuracy (\%) when varying the number of generated samples per category on the Office-Home dataset with ResNet-50 as the backbone.}
\label{officehome_number}
\setlength{\tabcolsep}{1mm}{
\begin{tabular}{ccccccccccccc @{\hskip 0.05in} c}
\toprule Number & Ar$\rightarrow$Cl   & Ar$\rightarrow$Pr  & Ar$\rightarrow$Rw & Cl$\rightarrow$Ar & Cl$\rightarrow$Pr & Cl$\rightarrow$Rw & Pr$\rightarrow$Ar & Pr$\rightarrow$Cl & Pr$\rightarrow$Rw & Rw$\rightarrow$Ar & Rw$\rightarrow$Cl & Rw$\rightarrow$Pr & Average \\
\midrule
0  & 57.2 & 77.2 & 82.0 & 66.5 & 77.2 & 76.7 & 62.3 & 56.3& 82.2 &  75.6 & 63.9 & 85.4 & 71.9  \\
50  & 59.7 & 78.4 & 83.0 & 68.6 & 81.5 & 79.8 & 64.7 & 57.4 & 83.5 & 75.4 & 64.7 & 86.5 & 73.6  \\
100  & 59.7 & 78.4 & 83.0 & 68.6 & 81.5 & 79.8 & 64.7 & 57.4 & 83.5 & 75.4 & 64.7 & 86.5 & 73.6  \\
200  & 59.9 & 78.8 & 83.2 & 67.8 & 81.1 & 80.6 & 63.9 & 57.9 & 83.2 & 75.9 & 64.2 & 86.1 & 73.6    \\
300  & 60.3 & 78.9 & 83.4 & 69.2 & 81.6 & 80.7 & 64.6 & 57.1 & 83.5 & 75.4 & 64.9 & 86.5 & 73.8  \\
400  & 60.6 & 79.1 & 83.5 & 68.9 & 81.7 & 80.9 & 63.7 & 58.0 & 83.7 & 75.2 & 64.7 & 86.4 & 73.8   \\
\bottomrule
\end{tabular}}
\end{table*}

\begin{table*}[!tbph]\small
\centering
\caption{Accuracy (\%) when varying the number of generated samples per category on the VisDA-2017 dataset with ResNet-101 as the backbone. }
\label{visda_number}
\setlength{\tabcolsep}{1.5mm}{
\begin{tabular}{ccccccccccccc @{\hskip 0.2in} c}
\toprule Number & aero & bicycle & bus & car & horse & knife & motor & person & plant & skate & train & truck & mean \\
\midrule
0 & 95.3 & 84.9 & 75.2 & 66.3 & 93.0 & 93.8 & 88.4 & 79.4 & 90.4 & 92.5 & 83.8 & 57.7 & 83.4 \\
500  & 95.9 & 83.9 & 80.2 & 69.9 & 94.1 & 93.0 & 90.3 & 81.5 & 91.0 & 94.3 & 86.4 & 56.5 & 84.7 \\
1000 & 96.1 & 82.5 & 81.6 & 72.3 & 94.9 & 97.3 & 90.4 & 80.8 & 92.9 & 92.4 & 85.8 & 56.9 & 85.3 \\
2000 & 96.6 & 86.1 & 82.2 & 68.8 & 95.4 & 96.3 & 90.8 & 82.5 & 92.7 & 93.3 & 87.9 & 60.5 & 86.1 \\
3000 & 96.5 & 85.2 & 83.0 & 74.7 &95.0 & 97.3 & 89.5 & 82.1 & 93.2 & 93.5 & 87.0 & 54.7 & 86.0 \\
4000 & 96.1 & 84.1 & 83.2 & 73.5 & 93.4 & 97.1 & 90.7 & 83.0 & 93.8 & 93.6 & 87.8 & 59.5 & 86.3 \\
\bottomrule
\end{tabular}}
\end{table*}

\section{Conclusion}
In this paper, we propose the DTS method for UDA. The DTS method is to generate high-quality samples for the target domain via CDPM. The generated target samples with pseudo labels are incorporated into source samples to form an augmented source domain, which could have a smaller domain discrepancy to the target domain than that between the original source and target domains.
In this way, the DTS method could help improve the transfer performance.
The DTS method is easy to implement and can be embedded into any UDA algorithm.
Extensive experiments demonstrate the effectiveness of the proposed DTS method.
In our future work, we are interested in applying the DTS method to other settings in transfer learning.


\section*{Acknowledgements}

This work is supported by NSFC key grant 62136005, NSFC general grant 62076118, and Shenzhen fundamental research program JCYJ20210324105000003.

\bibliographystyle{ieee_fullname}
\bibliography{DTS}

\begin{thebibliography}{10}\itemsep=-1pt

\bibitem{ben2010theory}
Shai Ben-David, John Blitzer, Koby Crammer, Alex Kulesza, Fernando Pereira, and
  Jennifer~Wortman Vaughan.
\newblock A theory of learning from different domains.
\newblock {\em Mach. learn.}, 79(1):151--175, May 2010.

\bibitem{chen2022reusing}
Lin Chen, Huaian Chen, Zhixiang Wei, Xin Jin, Xiao Tan, Yi Jin, and Enhong
  Chen.
\newblock Reusing the task-specific classifier as a discriminator:
  Discriminator-free adversarial domain adaptation.
\newblock In {\em Proceedings of the IEEE/CVF Conference on Computer Vision and
  Pattern Recognition}, pages 7181--7190, 2022.

\bibitem{cui2020gradually}
Shuhao Cui, Shuhui Wang, Junbao Zhuo, Chi Su, Qingming Huang, and Qi Tian.
\newblock Gradually vanishing bridge for adversarial domain adaptation.
\newblock In {\em Proceedings of the IEEE/CVF conference on computer vision and
  pattern recognition}, pages 12455--12464, 2020.

\bibitem{dhariwal2021diffusion}
Prafulla Dhariwal and Alexander Nichol.
\newblock Diffusion models beat gans on image synthesis.
\newblock {\em Advances in Neural Information Processing Systems},
  34:8780--8794, 2021.

\bibitem{foret2020sharpness}
Pierre Foret, Ariel Kleiner, Hossein Mobahi, and Behnam Neyshabur.
\newblock Sharpness-aware minimization for efficiently improving
  generalization.
\newblock {\em arXiv preprint arXiv:2010.01412}, 2020.

\bibitem{ganin2015unsupervised}
Yaroslav Ganin and Victor Lempitsky.
\newblock Unsupervised domain adaptation by backpropagation.
\newblock In {\em International conference on machine learning}, pages
  1180--1189. PMLR, 2015.

\bibitem{ganin2016domain}
Yaroslav Ganin, Evgeniya Ustinova, Hana Ajakan, Pascal Germain, Hugo
  Larochelle, Fran{\c{c}}ois Laviolette, Mario Marchand, and Victor Lempitsky.
\newblock Domain-adversarial training of neural networks.
\newblock {\em The journal of machine learning research}, 17(1):2096--2030,
  2016.

\bibitem{gao2022middlegan}
Ye Gao, Zhendong Chu, Hongning Wang, and John Stankovic.
\newblock Middlegan: Generate domain agnostic samples for unsupervised domain
  adaptation.
\newblock {\em arXiv preprint arXiv:2211.03144}, 2022.

\bibitem{goodfellow2014generative}
Ian Goodfellow, Jean Pouget-Abadie, Mehdi Mirza, Bing Xu, David Warde-Farley,
  Sherjil Ozair, Aaron Courville, and Yoshua Bengio.
\newblock Generative adversarial nets.
\newblock In {\em Proc. Adv. Neural Inf. Process. Syst.}, volume~27, pages
  2672--2680, 2014.

\bibitem{goodfellow2014explaining}
Ian~J Goodfellow, Jonathon Shlens, and Christian Szegedy.
\newblock Explaining and harnessing adversarial examples.
\newblock {\em arXiv preprint arXiv:1412.6572}, 2014.

\bibitem{goyal2022test}
Sachin Goyal, Mingjie Sun, Aditi Raghunathan, and J~Zico Kolter.
\newblock Test time adaptation via conjugate pseudo-labels.
\newblock In Alice~H. Oh, Alekh Agarwal, Danielle Belgrave, and Kyunghyun Cho,
  editors, {\em Advances in Neural Information Processing Systems}, 2022.

\bibitem{gretton2012kernel}
Arthur Gretton, Karsten~M Borgwardt, Malte~J Rasch, Bernhard Sch{\"o}lkopf, and
  Alexander Smola.
\newblock A kernel two-sample test.
\newblock {\em The Journal of Machine Learning Research}, 13(1):723--773, 2012.

\bibitem{he2016deep}
Kaiming He, Xiangyu Zhang, Shaoqing Ren, and Jian Sun.
\newblock Deep residual learning for image recognition.
\newblock In {\em Proceedings of the IEEE conference on computer vision and
  pattern recognition}, pages 770--778, 2016.

\bibitem{ho2020denoising}
Jonathan Ho, Ajay Jain, and Pieter Abbeel.
\newblock Denoising diffusion probabilistic models.
\newblock {\em Advances in Neural Information Processing Systems},
  33:6840--6851, 2020.

\bibitem{hoffman2018cycada}
Judy Hoffman, Eric Tzeng, Taesung Park, Jun-Yan Zhu, Phillip Isola, Kate
  Saenko, Alexei Efros, and Trevor Darrell.
\newblock Cycada: Cycle-consistent adversarial domain adaptation.
\newblock In {\em International conference on machine learning}, pages
  1989--1998. Pmlr, 2018.

\bibitem{huang2021variational}
Chin-Wei Huang, Jae~Hyun Lim, and Aaron~C Courville.
\newblock A variational perspective on diffusion-based generative models and
  score matching.
\newblock {\em Advances in Neural Information Processing Systems},
  34:22863--22876, 2021.

\bibitem{jiang2022transferability}
Junguang Jiang, Yang Shu, Jianmin Wang, and Mingsheng Long.
\newblock Transferability in deep learning: A survey.
\newblock {\em arXiv preprint arXiv:2201.05867}, 2022.

\bibitem{jin2020minimum}
Ying Jin, Ximei Wang, Mingsheng Long, and Jianmin Wang.
\newblock Minimum class confusion for versatile domain adaptation.
\newblock In {\em Computer Vision--ECCV 2020: 16th European Conference,
  Glasgow, UK, August 23--28, 2020, Proceedings, Part XXI 16}, pages 464--480.
  Springer, 2020.

\bibitem{long2015learning}
Mingsheng Long, Yue Cao, Jianmin Wang, and Michael Jordan.
\newblock Learning transferable features with deep adaptation networks.
\newblock In {\em International conference on machine learning}, pages 97--105.
  PMLR, 2015.

\bibitem{long2018conditional}
Mingsheng Long, Zhangjie Cao, Jianmin Wang, and Michael~I Jordan.
\newblock Conditional adversarial domain adaptation.
\newblock {\em Advances in neural information processing systems}, 31, 2018.

\bibitem{lu2022dpmsolver}
Cheng Lu, Yuhao Zhou, Fan Bao, Jianfei Chen, Chongxuan Li, and Jun Zhu.
\newblock {DPM}-solver: A fast {ODE} solver for diffusion probabilistic model
  sampling in around 10 steps.
\newblock In Alice~H. Oh, Alekh Agarwal, Danielle Belgrave, and Kyunghyun Cho,
  editors, {\em Advances in Neural Information Processing Systems}, 2022.

\bibitem{lu2022dpm}
Cheng Lu, Yuhao Zhou, Fan Bao, Jianfei Chen, Chongxuan Li, and Jun Zhu.
\newblock Dpm-solver++: Fast solver for guided sampling of diffusion
  probabilistic models.
\newblock {\em arXiv preprint arXiv:2211.01095}, 2022.

\bibitem{murphy2023probabilistic}
Kevin~P Murphy.
\newblock {\em Probabilistic machine learning: Advanced topics}.
\newblock MIT Press, 2023.

\bibitem{na2021fixbi}
Jaemin Na, Heechul Jung, Hyung~Jin Chang, and Wonjun Hwang.
\newblock Fixbi: Bridging domain spaces for unsupervised domain adaptation.
\newblock In {\em Proceedings of the IEEE/CVF Conference on Computer Vision and
  Pattern Recognition}, pages 1094--1103, 2021.

\bibitem{nichol2021improved}
Alexander~Quinn Nichol and Prafulla Dhariwal.
\newblock Improved denoising diffusion probabilistic models.
\newblock In {\em International Conference on Machine Learning}, pages
  8162--8171. PMLR, 2021.

\bibitem{ouyang2021maximum}
Liwen Ouyang and Aaron Key.
\newblock Maximum mean discrepancy for generalization in the presence of
  distribution and missingness shift.
\newblock In {\em NeurIPS 2021 Workshop on Distribution Shifts: Connecting
  Methods and Applications}, 2021.

\bibitem{peng2017visda}
Xingchao Peng, Ben Usman, Neela Kaushik, Judy Hoffman, Dequan Wang, and Kate
  Saenko.
\newblock Visda: The visual domain adaptation challenge.
\newblock {\em arXiv preprint arXiv:1710.06924}, 2017.

\bibitem{rangwani2022closer}
Harsh Rangwani, Sumukh~K Aithal, Mayank Mishra, Arihant Jain, and
  Venkatesh~Babu Radhakrishnan.
\newblock A closer look at smoothness in domain adversarial training.
\newblock In {\em International Conference on Machine Learning}, pages
  18378--18399. PMLR, 2022.

\bibitem{ronneberger2015u}
Olaf Ronneberger, Philipp Fischer, and Thomas Brox.
\newblock U-net: Convolutional networks for biomedical image segmentation.
\newblock In {\em Medical Image Computing and Computer-Assisted
  Intervention--MICCAI 2015: 18th International Conference, Munich, Germany,
  October 5-9, 2015, Proceedings, Part III 18}, pages 234--241. Springer, 2015.

\bibitem{saenko2010adapting}
Kate Saenko, Brian Kulis, Mario Fritz, and Trevor Darrell.
\newblock Adapting visual category models to new domains.
\newblock In {\em Computer Vision--ECCV 2010: 11th European Conference on
  Computer Vision, Heraklion, Crete, Greece, September 5-11, 2010, Proceedings,
  Part IV 11}, pages 213--226. Springer, 2010.

\bibitem{shen2018wasserstein}
Jian Shen, Yanru Qu, Weinan Zhang, and Yong Yu.
\newblock Wasserstein distance guided representation learning for domain
  adaptation.
\newblock In {\em Proceedings of the AAAI Conference on Artificial
  Intelligence}, volume~32, 2018.

\bibitem{Shi_2021_ICCV}
Lei Shi, Yifan Zhang, Jian Cheng, and Hanqing Lu.
\newblock Adasgn: Adapting joint number and model size for efficient
  skeleton-based action recognition.
\newblock In {\em Proceedings of the IEEE/CVF International Conference on
  Computer Vision (ICCV)}, pages 13413--13422, October 2021.

\bibitem{sohl2015deep}
Jascha Sohl-Dickstein, Eric Weiss, Niru Maheswaranathan, and Surya Ganguli.
\newblock Deep unsupervised learning using nonequilibrium thermodynamics.
\newblock In {\em International Conference on Machine Learning}, pages
  2256--2265. PMLR, 2015.

\bibitem{song2020denoising}
Jiaming Song, Chenlin Meng, and Stefano Ermon.
\newblock Denoising diffusion implicit models.
\newblock {\em arXiv preprint arXiv:2010.02502}, 2020.

\bibitem{su2018variational}
Jianlin Su.
\newblock Variational inference: A unified framework of generative models and
  some revelations.
\newblock {\em arXiv preprint arXiv:1807.05936}, 2018.

\bibitem{subramanian2022generalization}
Vignesh Subramanian, Rahul Arya, and Anant Sahai.
\newblock Generalization for multiclass classification with overparameterized
  linear models.
\newblock In Alice~H. Oh, Alekh Agarwal, Danielle Belgrave, and Kyunghyun Cho,
  editors, {\em Advances in Neural Information Processing Systems}, 2022.

\bibitem{sun2022prior}
Tao Sun, Cheng Lu, and Haibin Ling.
\newblock Prior knowledge guided unsupervised domain adaptation.
\newblock In {\em Computer Vision--ECCV 2022: 17th European Conference, Tel
  Aviv, Israel, October 23--27, 2022, Proceedings, Part XXXIII}, pages
  639--655. Springer, 2022.

\bibitem{van2008visualizing}
Laurens Van~der Maaten and Geoffrey Hinton.
\newblock Visualizing data using t-sne.
\newblock {\em Journal of machine learning research}, 9(11), 2008.

\bibitem{vapnik1999nature}
Vladimir Vapnik.
\newblock {\em The nature of statistical learning theory}.
\newblock Springer science \& business media, 1999.

\bibitem{venkateswara2017deep}
Hemanth Venkateswara, Jose Eusebio, Shayok Chakraborty, and Sethuraman
  Panchanathan.
\newblock Deep hashing network for unsupervised domain adaptation.
\newblock In {\em Proceedings of the IEEE conference on computer vision and
  pattern recognition}, pages 5018--5027, 2017.

\bibitem{wang2023energyinspired}
Ze Wang, Jiang Wang, Zicheng Liu, and Qiang Qiu.
\newblock Energy-inspired self-supervised pretraining for vision models.
\newblock In {\em International Conference on Learning Representations}, 2023.

\bibitem{AFN}
Ruijia Xu, Guanbin Li, Jihan Yang, and Liang Lin.
\newblock Larger norm more transferable: An adaptive feature norm approach for
  unsupervised domain adaptation.
\newblock In {\em ICCV}, 2019.

\bibitem{yang2020bi}
Guanglei Yang, Haifeng Xia, Mingli Ding, and Zhengming Ding.
\newblock Bi-directional generation for unsupervised domain adaptation.
\newblock In {\em Proceedings of the AAAI conference on artificial
  intelligence}, volume~34, pages 6615--6622, 2020.

\bibitem{yang2022diffusion}
Ling Yang, Zhilong Zhang, Yang Song, Shenda Hong, Runsheng Xu, Yue Zhao,
  Yingxia Shao, Wentao Zhang, Bin Cui, and Ming-Hsuan Yang.
\newblock Diffusion models: A comprehensive survey of methods and applications.
\newblock {\em arXiv preprint arXiv:2209.00796}, 2022.

\bibitem{yang2020transfer}
Qiang Yang, Yu Zhang, Wenyuan Dai, and Sinno~Jialin Pan.
\newblock {\em Transfer learning}.
\newblock Cambridge University Press, 2020.

\bibitem{NEURIPS2021_ae0909a3}
Werner Zellinger, Natalia Shepeleva, Marius-Constantin Dinu, Hamid
  Eghbal-zadeh, Hoan~Duc Nguyen, Bernhard Nessler, Sergei Pereverzyev, and
  Bernhard~A. Moser.
\newblock The balancing principle for parameter choice in distance-regularized
  domain adaptation.
\newblock In M. Ranzato, A. Beygelzimer, Y. Dauphin, P.S. Liang, and J.~Wortman
  Vaughan, editors, {\em Advances in Neural Information Processing Systems},
  volume~34, pages 20798--20811. Curran Associates, Inc., 2021.

\bibitem{zhang2022spectral}
Jingyi Zhang, Jiaxing Huang, Zichen Tian, and Shijian Lu.
\newblock Spectral unsupervised domain adaptation for visual recognition.
\newblock In {\em Proceedings of the IEEE/CVF Conference on Computer Vision and
  Pattern Recognition}, pages 9829--9840, 2022.

\bibitem{MDD}
Yuchen Zhang, Tianle Liu, Mingsheng Long, and Michael Jordan.
\newblock Bridging theory and algorithm for domain adaptation.
\newblock In {\em ICML}, 2019.

\bibitem{zhang2023free}
YiFan Zhang, Xue Wang, Jian Liang, Zhang Zhang, Liang Wang, Rong Jin, and
  Tieniu Tan.
\newblock Free lunch for domain adversarial training: Environment label
  smoothing.
\newblock In {\em International Conference on Learning Representations}, 2023.

\bibitem{zhao2020review}
Sicheng Zhao, Xiangyu Yue, Shanghang Zhang, Bo Li, Han Zhao, Bichen Wu, Ravi
  Krishna, Joseph~E Gonzalez, Alberto~L Sangiovanni-Vincentelli, Sanjit~A
  Seshia, et~al.
\newblock A review of single-source deep unsupervised visual domain adaptation.
\newblock {\em IEEE Transactions on Neural Networks and Learning Systems},
  33(2):473--493, 2020.

\bibitem{zhu2017unpaired}
Jun-Yan Zhu, Taesung Park, Phillip Isola, and Alexei~A Efros.
\newblock Unpaired image-to-image translation using cycle-consistent
  adversarial networks.
\newblock In {\em Proceedings of the IEEE international conference on computer
  vision}, pages 2223--2232, 2017.

\bibitem{zhu2020deep}
Yongchun Zhu, Fuzhen Zhuang, Jindong Wang, Guolin Ke, Jingwu Chen, Jiang Bian,
  Hui Xiong, and Qing He.
\newblock Deep subdomain adaptation network for image classification.
\newblock {\em IEEE transactions on neural networks and learning systems},
  32(4):1713--1722, 2020.

\end{thebibliography}

\end{document}